\documentclass[12pt, hidelinks]{article}

\usepackage[utf8]{inputenc}
\usepackage{amsmath, amssymb, url, float, graphicx, float,booktabs}
\usepackage{algorithm, algpseudocode}
\usepackage{fullpage}
\usepackage[small,bf]{caption}

\newcommand{\normsq}[1]{\left\|{#1}\right\|_2^2}

\newcommand{\ones}{\mathbf 1}
\newcommand{\reals}{{\mbox{\bf R}}}

\newcommand{\diag}{\mathop{\bf diag}}

\newcommand{\prox}{\mathbf{prox}}

\newcommand{\Prob}{\mathop{\bf Prob}}

\DeclareMathOperator*{\argmin}{argmin}
\DeclareMathOperator*{\argmax}{argmax}

\newcommand{\eg}{{\it e.g.}}
\newcommand{\ie}{{\it i.e.}}

\newcommand{\BEAS}{\begin{eqnarray*}}
\newcommand{\EEAS}{\end{eqnarray*}}
\newcommand{\BEA}{\begin{eqnarray}}
\newcommand{\EEA}{\end{eqnarray}}
\newcommand{\BEQ}{\begin{equation}}
\newcommand{\EEQ}{\end{equation}}
\newcommand{\BIT}{\begin{itemize}}
\newcommand{\EIT}{\end{itemize}}

\title{Optimal Representative Sample Weighting}
\author{Shane Barratt \and Guillermo Angeris \and Stephen Boyd}
\date{\today}

\begin{document} 
\maketitle 
\begin{abstract}
We consider the problem of assigning weights
to a set of samples or data records,
with the goal of achieving a representative weighting,
which happens when certain sample averages of
the data are close to prescribed values.
We frame the problem of finding representative sample weights
as an optimization problem,
which in many cases is convex and can be efficiently solved.
Our formulation includes as a special case the selection
of a fixed number of the samples, with equal weights, \ie,
the problem of selecting a smaller representative
subset of the samples.
While this problem is combinatorial and not convex, heuristic methods based
on convex optimization seem to perform very well.
We describe \verb|rsw|, an open-source implementation of the ideas
described in this paper, and apply it to a skewed sample of the CDC BRFSS dataset.
\end{abstract}

\section{Introduction}
We consider a setting where we have a set of data samples that were not uniformly
sampled from a population, or where they were sampled from a different 
population than the one from which we wish to draw some conclusions.
A common approach is to assign \emph{weights} to the samples,
so the resulting weighted distribution is \emph{representative} of the 
population we wish to study.
Here representative means that with the weights, certain expected 
values or probabilities match or are close to known values for the 
population we wish to study.

A a very simple example, consider a data set where each sample is
associated with a person.  Our data set is 70\% female, whereas 
we'd like to draw conclusions about a population that is 50\% female.
A simple solution is to down-weight the female samples, and up-weight the
male samples in our data set, so the weighted fraction of females is 50\%.
As a more sophisticated example, suppose we have multiple groups,
for example various combinations of sex, age group, income level, and education,
and our goal is to find weights for our samples so the fractions
of all these groups matches or approximates known fractions in
the population we wish to study.  In this case, there will be many possible
assignments of weights that match the given fractions, and we need to
choose a reasonable one.  One approach is to maximize the entropy of the 
weights, subject to matching the given fractions.  This problem generally
does not have an analytical solution but it can be solved by 
various methods, for example,
\emph{raking} or \emph{iterative proportional fitting}~\cite[\S7]{lumley2011complex},
or the methods described in this paper.

We formulate a general method for choosing weights that balances
two objectives: making the weighted sample representative, \ie,
matching some known fractions or other statistics, and having weights
with desirable properties, such as being close to uniform.
We pose this as an optimization problem.  In many practical cases,
the problem is convex, and so can be solved efficiently~\cite{cvxbook}; in others,
the problem can be expressed as a mixed-integer convex optimization
problem, which in general is hard to solve exactly, but for which
simple and effective heuristics can be developed.

Our method allows for a variety of measures of how representative
a weighted data set is, and a variety of other desirable properties for
the weights themselves.  Simple examples of measures of representativeness 
include allowing the 
user to specify ranges for fractions or expected values, instead of 
insisting that the values are matched precisely.  For example, we might
specify that the weighted sample should have a fraction of females 
between 49\% and 51\%, instead of insisting that it should be exactly 50\%.
As a simple example of desirable properties for the weights, we can specify
minimum and maximum allowable values for the weights, \eg, we can require
that the weights range between one half and twice the uniform weighting,
so no sample is down- or up-weighted by more than a factor of two.

One particularly interesting constraint we can impose on the weights
is that they are all zero, except for a specified number $k$ of them,
and those must have the same weight, $1/k$.  The weighted data set 
is then nothing more than a subset of $k$ of our original samples, with
the usual uniform weights.  This means that we are selecting a set of $k$
samples from our original data set that is representative.
This can be useful when we need to carry out some expensive further testing of
some of the samples, and we wish to choose a subset of them that 
is representative.

In this paper we present a general framework for choosing weights,
and a general algorithm for solving the associated 
optimization problems.  Our method, which is based on an operator splitting 
method called ADMM (alternating directions method of multipliers),
solves the problem exactly when it is convex, and is a good heuristic 
when the problem is not convex.
We have implemented the method as an open source software package
in Python, called \texttt{rsw} (for representative sample weights).
It handles a wide variety of weight selection problems, and scales to 
very large problems with hundreds of thousands or millions of samples,
making it suitable for almost all practical surveys.

\paragraph{Related work.} Survey weighting
is a well-studied problem in statistics (see, \eg,~\cite{valliant2013practical}
for an overview).
The simplest technique for survey weighting is post-stratification~\cite{neyman1934two}.
When the samples can be divided into a finite number of groups,
and we want the fraction of samples in each group to be equal to some given fraction,
post-stratification adjusts the weights of each group using
a simple formula so that the fraction of samples in the
weighted groups match the given fractions~\cite{holt1979post}.
For example, if we had 4 female samples and 6 male samples,
and we wanted half to be female, we would give each female
sample a weight of 0.125 and each male sample a weight of 0.083.
Post-stratification, when applied to multiple attributes, requires
the groups to be constructed as each combination of the attributes,
which might lead to many groups and be undesirable~\cite[\S7.3]{lumley2011complex}.

An alternative to post-stratification, when there are multiple attributes,
is to match the marginal distribution for each attribute instead of the
full joint distribution, while maximizing the entropy of the sample weights.
This problem is exactly the maximum entropy problem described in \S\ref{sec:max-entropy}.
The most common iterative algorithm used in practice is raking (also known as iterative proportional
fitting or rim weighting)~\cite{yule1912methods,kruithof1937telefoonverkeersrekening,deming1940least},
which works by cycling through each attribute and performing post-stratification in order
to gradually update the sample weights.
Raking is the standard weighting method used by many public pollsters, according
to the Pew Research Center~\cite{pew}.
It was observed that a version of raking can be seen as coordinate
ascent on the dual of the maximum entropy problem, which is guaranteed to converge
\cite{deville1993generalized,teh2003improving} (see appendix~\ref{app:coordinate-ascent}).
Other optimization-based formulations of survey weighting can be found in
\cite{fu2019cvxr,she2019iterative}.

There are a number of generalizations to the raking procedure, that are each for
different regularizers~\cite[\S3]{deville1993generalized}.
Some of these are linear weighting~\cite{bethlehem1987linear}
(which uses the Pearson $\mathcal X^2$-divergence between
the weights and the prior on the weights as the regularizer), logit weighting, and truncated
linear weighting.
Each of these has a corresponding generalized raking procedure,
yet all of these generalizations are instances of our formulation,
and can be easily solved using our ADMM algorithm even when
the loss or regularizer is nonsmooth.
Another technique is logistic regression weighting,
which regresses the probability of sampling each sample
on the sample attributes, and uses the reciprocal of the predicted
sampling probabilities as weights~\cite{lepkowski1989weighting,iannacchione1991response}.
We note that various heuristic techniques
have been proposed for constraining the weights in the raking algorithm~\cite[\S3]{kalton2003weighting},
but, in our formulation, these can be easily expressed as constraints
included in the regularizer.
To our best knowledge, the representative selection
problem, described in \S\ref{sec:representative_selection},
has not appeared in the literature.
The closest problem is representative subset selection~\cite{daszykowski2002representative},
which tries to select a set of samples that are as different as possible from one
another.

Once sample weights have been found, they can be used in a
Horvitz–Thompson estimator to estimate various quantities~\cite{horvitz1952generalization}.
They can also be used in down-stream statistical analysis, \eg, linear regression~\cite{lavallee2015we}.
We discuss these and some other applications for weights in \S\ref{s-applications}.
For a discussion of whether weighting is appropriate or not for a particular
dataset or goal, see, \eg,~\cite{kish1992weighting}.
(Our focus is on finding weights, and not whether or not using them is appropriate.)

Another important aspect of survey weighting is non-response weighting (also known as
propensity weighting or direct standardization).
Non-response weighting is the first part of the weighting process,
which adjusts the weight of each sample based on the probability that a sample
with those attributes would not respond to the survey~\cite[\S9]{lumley2011complex}.
The most common technique is to construct a logistic regression model,
which regresses the response probability on the sample attributes~\cite{heckman1976common}.
We assume that non-response weighting has already been performed, or that there
is no non-response bias in the dataset.

\paragraph{Outline.}
In \S\ref{sec:rep_samp_weighting} we describe the setting
of sample weighting and the optimization problem of
finding representative sample weights.
In \S\ref{sec:examples} we provide a number of examples of
the representative sample weighting problem.
In \S\ref{sec:solution_method} we give a solution method
based on ADMM.
In \S\ref{s-applications} we describe a number of applications
once sample weights have been computed.
In \S\ref{s-experiments} we demonstrate our method on annual
survey data from the U.S.\ CDC.
We conclude in \S\ref{s-ext-var} with extensions and variations.

Appendix~\ref{app:coordinate-ascent} gives a proof that raking
is equivalent to performing block coordinate ascent
on the dual variables for a particular instance of the representative sample
weighting problem, and appendix~\ref{sec:desired_tables} gives a table of
desired expectations for the provided examples.

\section{Representative sample weighting}
\label{sec:rep_samp_weighting}

In this section we describe the general problem of representative sample
weighting.

\paragraph{Samples and weights.}
We are given $n$ samples
or data records $x_1,\ldots,x_n\in\mathcal X$.
For our purposes the feature values $x_i$ and the feature set 
$\mathcal X$ do not matter; they will enter our problem 
in a very specific way, described below.
We will assign a weight $w_i\in\reals_+$ to each
sample $x_i$, $i=1,\ldots,n$,
where $\ones^Tw=1$ and $\ones$ is the all-ones vector.
The sample weights induce a distribution on $\{x_1,\ldots, x_n\}$,
with $\Prob(x_i) = w_i$, $i=1,\ldots,n$.
Our goal is to choose these sample weights.

\paragraph{Expected values of functions.}
We are given real-valued functions 
$F_j:\mathcal X\to\reals$, $j=1,\ldots,m$.
We let $f \in \reals^m$ denote the expected values of these functions
under the induced distribution,
\[
f_j = \sum_{i=1}^n w_i F_j(x_i), \quad j=1,\ldots, m.
\]
We express this as $f=Fw$, where $F\in \reals^{m \times n}$ is the matrix
with entries $F_{ij} = F_i(x_j)$.
In the sequel we will only need this matrix $F$; we will not otherwise
need the particular values of $x_i$ or the functions $F_j$.

\paragraph{Probabilities.}
We note that when the function $F_j$ is the $\{0,1\}$ indicator 
function of a set $S \subseteq \mathcal X$, 
$f_j$ is the probability of $S$ under the induced distribution.
As an elementary example of this, $S$ could be the set of examples
that are female (which is encoded in $x$),
in which case $f_j$ is the probability, under the 
weights $w$, that the sample is female.

As an extension of this, suppose $F_1, \ldots, F_m$ are the $\{0,1\}$
indicator functions of the sets $S_1, \ldots, S_m \subseteq \mathcal X$, and each
example is in exactly one of these sets (\ie, they are a partition).
In this case $f=Fw$ is a distribution on $\{1,\ldots, m\}$,
\ie, $f \geq 0$ (elementwise) and $\ones^Tf=1$.
As an elementary example, these could be a partition of samples
into female and male, and a number of age groups; $f_j$ is then
the probability under the weighted distribution that an example
is in group $j$, which is some combination of sex and age group.

\paragraph{Desired expected values and loss function.}
Our task is to choose the weights $w \in \reals_+^n$
so that this induced distribution
is \emph{representative}, which means that the expected values of 
the functions of $x_i$ are near some target or desired values,
denoted $f^\mathrm{des}\in\reals^m$.
To get a numerical measure of how representative the induced
distribution is, we introduce a 
loss function $\ell:\reals^m\times\reals^m\to\reals\cup\{+\infty\}$,
where $\ell(f,f^\mathrm{des})$
measures our dissatisfaction with a particular
induced expected value $f$ and desired expected value $f^\mathrm{des}$. 
We use infinite values of $\ell$ to denote constraints on
$f$ and $f^\mathrm{des}$.  For example, to constrain $f=f^\mathrm{des}$,
we let $\ell(f,f^\mathrm{des})=+\infty$ for $f\neq f^\mathrm{des}$,
and $\ell(f,f^\mathrm{des})=0$ for $f = f^\mathrm{des}$.
Other common losses include the sum squares loss
$\|f-f^\mathrm{des}\|_2^2$, or the $\ell_1$ loss
$\|f-f^\mathrm{des}\|_1$.

When $f$ is a distribution on $\{1,\ldots, m\}$
(\ie, for all $w$ with $w\geq 0$, $\ones^T w=1$, we have
$Fw \geq 0$ and $\ones^TF w=1$),
a reasonable loss is the Kullback-Liebler (KL) divergence from $f^\mathrm{des}$,
a desired target distribution,
\[
\ell(f,f^\mathrm{des}) = 
\sum_{i=1}^m f_i \log (f_i / f^\mathrm{des}_i).
\]
Here we assume all entries of $f$ together form a distribution; if there
are multiple groups of entries in $f$ that form distributions we can use
the sum of the KL divergences from their associated target distributions.
As an example and special case suppose $f_i$ is the probability that $x \in S_i$,
where $S_i \subseteq \mathcal X$ are sets (not necessarily a partition),
for which we have target values $f^\mathrm{des}_i$.
Then $1-f_i$ is the probability that $x \not\in S_i$.  A reasonable loss 
is the sum of the KL divergence from these two probabilities and their associated
desired probabilities, \ie,
\[
\ell(f,f^\mathrm{des}) = \sum_{i=1}^m
\left(
f_i \log (f_i/f^\mathrm{des}_i) + (1-f_i) \log((1-f_i)/(1-f^\mathrm{des}_i))
\right).
\]

\paragraph{Regularizer.}
We also have preferences and requirements on the choice of 
weights, beyond $w \geq 0$ and $\ones^Tw =1$, unrelated to the 
data samples and functions.  We express these using a 
regularization function $r: \reals^n \to \reals \cup \{\infty\}$,
where we prefer smaller values of $r(w)$.
Again, we use infinite values of $r$ to denote constraints on $w$.
For example, to constrain $w \geq w^\mathrm{min}$ for $w^\mathrm{min} \in \reals_+^n$,
we let $r(w) = +\infty$ whenever $w < w^\mathrm{min}$.

An important example of a regularizer is the negative entropy,
with 
\[
r(w) = \sum_{i=1}^n w_i \log w_i,
\]
which is the same as the KL divergence from the distribution given by
$w$ and the uniform distribution $\ones/n$ on $\{1,\ldots, n\}$.
This regularizer expresses our desire that the weights have high
entropy, or small KL deviation from uniform.
A simple extension of this regularizer is the KL deviation from a 
(non-uniform) target distribution $w^\text{tar}$.

Regularizers can be combined, as in
\BEQ
\label{eq:limit}
r(w) = \left\{ \begin{array}{ll} 
\sum_{i=1}^n w_i \log w_i & (1/(\kappa n))\ones \leq w \leq (\kappa/n) \ones\\
\infty & \text{otherwise}, \end{array} \right.
\EEQ
where $\kappa>1$ is a given hyper-parameter.
This regularizer combines negative entropy with the constraint
that no sample is up- or down-weighted by a factor more than $\kappa$.

\paragraph{Representative sample weighting problem.}
We propose to choose weights as a solution of
the \emph{representative sample weighting} problem
\BEQ
\begin{array}{ll}
\text{minimize} & \ell(f, f^\mathrm{des}) + \lambda r(w)\\
\text{subject to} & f = Fw,\quad
w \geq 0, \quad \ones^Tw = 1,
\end{array}
\label{eq:representative_sample}
\EEQ
with variables $w$ and $f$, and positive hyper-parameter $\lambda$.
The objective is a weighted sum of the loss, which measures how
unrepresentative the induced distribution is, and the regularization,
which measures our displeasure with the set of weights.
The hyper-parameter $\lambda$ is used to control the trade off between 
these two objectives.

The representative sample weighting problem~(\ref{eq:representative_sample})
is specified by the loss function $\ell$ and desired expected 
values $f^\mathrm{des}$, the regularizer $r$, and the matrix $F$ which 
gives the function values $F_i(x_j)$. 

\paragraph{Convexity.}
When $\ell$ is convex in its first argument
and $r$ is convex,
problem \eqref{eq:representative_sample}
is a convex optimization problem, and so can be efficiently solved
\cite{cvxbook}.
Many interesting and useful instances of the representative
sample weighting problem are convex; a few are not.
The cases when the problem is not convex can be solved 
using global optimization methods; we will also recommend
some simple but effective heuristics for solving such problems
approximately.

\section{Examples}
\label{sec:examples}

In this section we give two
examples of the representative sample weighting problem.

\subsection{Maximum entropy weighting}
\label{sec:max-entropy}
The first case we consider is where we want
to find the maximum entropy sample weights
that exactly match the desired expected values.
This is a representative sample weighting problem,
with the loss and regularizer
\[
\ell(f,f^\mathrm{des}) = \begin{cases}
0 & f = f^\mathrm{des}, \\
+\infty & \text{otherwise},
\end{cases}
\qquad
r(w) = \sum_{i=1}^n w_i \log w_i.
\]
Since both the loss and regularizer are
convex, problem \eqref{eq:representative_sample}
is a convex optimization problem.
We call the (unique) solution to problem \eqref{eq:representative_sample}
with this loss and regularizer
the \emph{maximum entropy weighting}.
The maximum entropy weighting is the most random
or evenly spread out weighting that exactly
matches the desired expected values.

The maximum entropy weighting problem is convex and readily solved 
by many methods. The typical complexity of such methods is order $nm^2$,
\ie, linear in the number of samples $n$
and quadratic in the number of functions $m$.
(See~\cite[\S11.5]{cvxbook}.)
This complexity assumes that the matrix $F$ is treated as dense; by exploiting
sparsity in $F$, the complexity of solving the problem can be lower.

\paragraph{Simple example.}
As a very simple example, suppose $m=1$ and $F_1$ is the $\{0,1\}$ 
indicator function of the sample being female, so $f=Fw$ is the probability
of being female under the weighted distribution, and we take $f^\mathrm{des}=0.5$,
which means we seek weights for which the probability of female is 0.5.
(This problem has an elementary and obvious solution, where we weight all 
female samples the same way, and all non-female samples the same way.
But our interest is in more complex problems that do not have simple
analytical solutions.)

\paragraph{Variations.}
We have already mentioned several variations, such as adding bounds
on $w$, or replacing the negative entropy with the KL divergence from 
a target weighting.
As another extension, we can modify the loss to allow the expected function
values $f$ to be close to, but not exactly, their desired values.
As a simple version of this, we can require that
$f^\mathrm{min} \leq f \leq f^\mathrm{max}$, 
where $f^\mathrm{min}, f^\mathrm{max} \in \reals^m$ are given 
lower and upper limits (presumably with
$f^\mathrm{min} \leq f^\mathrm{des} \leq f^\mathrm{max}$).
This corresponds to the loss function
\[
\ell(f,f^\mathrm{des}) = \begin{cases}
0 & f^\mathrm{min} \leq f \leq f^\mathrm{max}, \\
+\infty & \text{otherwise},
\end{cases}
\]
the indicator function of the 
constraint $f^\mathrm{min} \leq f \leq f^\mathrm{max}$.

In our simple example above, for example, we might take
$f^\mathrm{min}=0.45$ and $f^\mathrm{max}=0.55$.  This means that we will
accept any weights for which the probability of female is between $0.45$
and $0.55$.
(This problem also has an elementary analytical solution for this example.)

\subsection{Representative selection}
\label{sec:representative_selection}
In the second example, we consider the problem of selecting $k<n$
samples, with equal weights, that are as representative as possible.
Thus, our regularizer has the form 
\BEQ
r(w) = \begin{cases}
0 & w \in \{0,1/k\}^n,\\
+\infty & \text{otherwise}.
\end{cases}
\label{eq:boolean_reg}
\EEQ
This regularizer is infinite unless we have $w_i = 1/k$ for $i \in \mathcal I$,
and $w_i = 0$ for $i \not \in \mathcal I$, where $\mathcal I \subset \{1,\ldots,
n\}$ and $|\mathcal I| = k$.  In this case the induced distribution 
is uniform on $x_i$ for $i \in \mathcal I$.

With this constraint, it is unlikely that we
can achieve $f=f^\mathrm{des}$ exactly, so we use a loss 
function such as
$\ell(f,f^\mathrm{des}) = \|f-f^\mathrm{des}\|_2^2$,
$\ell(f,f^\mathrm{des}) = \|f-f^\mathrm{des}\|_1$,
or the KL divergence to $f^\mathrm{des}$ if $f$ is a distribution.

We will refer to this problem as \emph{representative selection}.
Roughly speaking, the goal is to choose a subset of size $k$ from the $n$ 
original data points, such that this subset is as representative as possible.
The problem is interesting even when we take $f^\mathrm{des} = (1/n)F\ones$,
in which case the problem is to choose a subset of the original
samples, of size $k$,
that approximates the means of the functions on the original data set.

Representative selection is a combinatorial optimization problem of
selecting $k$ out of $n$ original data points, of which there are $n \choose k$
choices.
This problem can be difficult to solve exactly, 
since you can solve the 0-1 integer problem
by solving $n$ instances of the representative selection problem
\cite{karpReducibilityCombinatorialProblems1972}.
Representative selection can be reformulated as a mixed-integer convex program,
for which there exist modern solvers (\eg, ECOS~\cite{bib:Domahidi2013ecos},
GUROBI~\cite{gurobi}, and MOSEK~\cite[\S9]{aps2020mosek}) that
can often find global solutions of small problems in a reasonable amount of time.
For the representative selection problem,
we describe an ADMM algorithm in~\S\ref{sec:solution_method}
that can quickly find an approximate solution,
which appears to have good practical performance
and can be used when $n$ is very large.
We also note that the representative selection problem
is similar in nature to the sparse approximation problem,
which can be approximately solved by heuristics such as basis or
matching pursuit~\cite{chen2001atomic}.

\section{Applications}
\label{s-applications}

In this section we describe what to do once sample weights
have been computed.

\paragraph{Direct use of weights.}
The most common use of sample weights is to use
them as weights in each subsequent step of a
traditional machine learning or statistical analysis pipeline.
Most machine learning and statistical methods are able
to incorporate sample weights, by in some way modulating the
importance of each sample by its weight in the fitting process.
As a result, samples with higher weights affect the ultimate model
parameters more than those with lower weights.
As a simple example, suppose we have another function $g:\mathcal X\to\reals$
for which we want to compute its mean.
Our na\"ive estimate of the mean would be the sample average,
$\sum_{i=1}^n (1/n)g(x_i)$; using the sample weights, our estimate
of the mean is $\sum_{i=1}^n w_i g(x_i)$.
(This is the aforementioned Horwitz-Thompson estimator.)

As another example, suppose we have additional features
of each sample denoted $a_1,\ldots,a_n\in\reals^p$
and outcomes $b_1,\ldots,b_n\in\reals$ and we wish to fit
a linear regression model, \ie, $\theta^Ta_i \approx b_i$ for some
$\theta\in\reals^p$. With sample weights, the fitting problem becomes
\[
\begin{array}{ll}
\text{minimize} & \sum_{i=1}^n w_i (\theta^Ta_i - b_i)^2.
\end{array}
\]

\paragraph{Re-sampling.}
Another option is to use the sample weights to re-sample the data,
and then use the new set of samples for further analysis.
That is, we create a new set of $N$ samples $\tilde x_1,\ldots,\tilde x_{N}$,
using the distribution
\[
\Prob(\tilde x_i=x_i) = w_i, \quad i=1,\ldots,N. 
\]
For many machine learning and statistical analysis methods,
when $N$ becomes very large,
re-sampling the data converges to the approach above of using
the weights directly.
However, re-sampling can be useful when the machine learning
or statistical analysis method cannot incorporate weights.

\paragraph{Selection.}
The representative selection problem,
described in \S\ref{sec:representative_selection},
can be used to select
a subset of the samples
that are themselves representative.
Representative selection can be useful
when we want to select a few of the samples
to gather more data on, but it is infeasible
(or difficult, or expensive) to gather more data on all of the samples.
This can be useful in a two-stage sampling or study,
where one first gathers basic (low-cost) data on
a large number of samples, and then selects a representative subset
of those samples to include in the second (high-cost)
study~\cite[\S8]{lumley2011complex}.

For future reference we mention that a simple heuristic for representative 
selection comes directly from weighted sampling, described above.
We start by solving the maximum entropy weight selection problem, and then
draw $k$ samples from the associated weighted distribution.
This simple scheme can be used as an initialization for a heuristic method
for the problem.

\section{Solution method}
\label{sec:solution_method}

The problem \eqref{eq:representative_sample} can be solved by many methods,
depending on the properties of the loss and regularizer functions.
For example, in the maximum entropy problem, with equality constraints on
$f$ or with sum square loss, Newton's method can be 
effectively used~\cite[\S11]{cvxbook}.
More complex problems, with nondifferentiable loss and regularizer,
can be solved using, for example, interior-point methods,
after appropriate transformations~\cite[\S11.6]{cvxbook}.
It is very easy to express general convex representative sample weighting problems
using domain specific languages (DSLs) for convex optimization,
such as CVX~\cite{grant2008cvx,grant2014cvx},
YALMIP~\cite{lofberg2004yalmip}, CVXPY~\cite{diamond2016cvxpy,agrawal2018cvxpy}, Convex.jl
\cite{udell2014convexjl}, and CVXR~\cite{fu2019cvxr}.
Many of these packages are also able to handle mixed-integer convex programs,
\eg, the representative selection problem can be expressed in them.
These DSLs canonicalize the (human-readable) problem into a convex
or mixed-integer convex cone program and call a numerical solver,
such as OSQP~\cite{stellato2020osqp}, SCS~\cite{odonoghue2016conic},
ECOS~\cite{bib:Domahidi2013ecos}, MOSEK~\cite{bib:Domahidi2013ecos}, or
GUROBI~\cite{gurobi}.
However, as we will see in \S\ref{s-experiments},
these generic solvers can often be quite slow, taking two orders of magnitude
more time to solve the problem compared to the algorithm we describe below.
(The advantage of these DSLs is that representative sampling problems
can be specified in just a few lines of code.)

In this section we describe a simple and efficient algorithm
for solving problems of the form \eqref{eq:representative_sample},
including losses and regularizers that take on infinite values and are
nondifferentiable or nonconvex.  While it can at times be slower
to achieve high accuracy than an interior-point method,
high accuracy is not needed in practice in
the representative sample weighting problem.
Our algorithm uses the alternating direction
method of multipliers (ADMM; see~\cite{boydDistributedOptimizationStatistical2010}).
When the loss and regularizer are convex, and the problem 
is feasible, this algorithm is guaranteed
to converge to a globally optimal solution.
In the case when the loss or regularizer is not convex, our method is a heuristic
that will find an approximate solution; it has been observed that these
approximate solutions are often good enough for practical purposes,
and that the approximate solutions can be refined or polished using greedy
methods~\cite{diamond2018general}.

\paragraph{Problem rewriting.} In order to apply ADMM to
problem~\eqref{eq:representative_sample},
we introduce two new redundant variables,
resulting in the problem
\begin{equation}\label{eq:admm-rewriting}
\begin{array}{ll}
\text{minimize} & \ell(f, f^\mathrm{des}) + \lambda r(\tilde w)\\
\text{subject to} & f = Fw\\
& \tilde w = w = \bar w\\
& \bar w \geq 0, \quad \ones^T\bar w = 1,
\end{array}
\end{equation}
with variables $w,\tilde w, \bar w\in\reals^n$ and $f\in\reals^m$.
The problem data is still the data matrix
$F \in \reals^{m \times n}$ and the desired values $f^\mathrm{des} \in \reals^m$.
Problem \eqref{eq:admm-rewriting} is said to be in
\emph{graph form}, and many algorithms exist
that can solve such problems, \eg, ADMM~\cite{parikh2014block}
and POGS~\cite{fougner2018parameter}.

In~(\ref{eq:admm-rewriting}) we replicate the variable $w$ into 
three versions, which must all be equal.
Each of these versions of $w$ handles one aspect of the problem: $\bar w$ 
is constrained to be a probability distribution; $\tilde w$ appears in
the regularizer, and $w$ appears in the loss.
The technique of replicating a variable and insisting that the different
versions be consistent is sometimes called
\emph{consensus optimization} 
\cite[\S7.1]{boydDistributedOptimizationStatistical2010}.
By itself, this consensus form is useless. It becomes useful when combined
with an operator splitting method that handles the different versions of $w$
separately.
We now describe one such operator splitting method, ADMM.

\paragraph{ADMM steps.} Using the splitting $(f, \tilde w, \bar w)$ and $w$,
the corresponding ADMM steps (see, \eg,~\cite[\S4.4]{parikhProximalAlgorithms2014})
are
\begin{equation}\label{eq:admm-steps}
\begin{aligned}
	f^{k+1} &= \prox_{(1/\rho)\ell(\cdot, f^\mathrm{des})} \left(Fw^{k} - y^k\right),\\
	\tilde w^{k+1} &= \prox_{(\lambda/\rho) r}\left(w^k + z^k\right),\\
	\bar w^{k+1} &= \Pi\left(w^k + u^k\right),\\
	w^{k+1} &= \argmin_{w}\left(\normsq{f^{k+1} - Fw + y^k} + \normsq{\tilde w^{k+1} - w + z^k} + \normsq{\bar w^{k+1} - w + u^k}\right),\\
	y^{k+1} &= y^k + f^{k+1} - Fw^{k+1},\\
	z^{k+1} &= z^{k} + w^{k+1} - \tilde w^{k+1},\\
	u^{k+1} &= u^{k} + w^{k+1} - \bar w^{k+1},
\end{aligned}
\end{equation}
where $y\in\reals^m$ and $z, u\in\reals^n$ are the (scaled) dual variables
for the constraints,
and $\rho > 0$ is a given penalty parameter.
Here, $\prox_g$ is the proximal operator of the function $g$
(see~\cite[\S1.1]{parikhProximalAlgorithms2014}) and often has closed-form solutions
for many convex (and some nonconvex) functions that appear in practice.
The function $\Pi$ is the projection onto the probability simplex
$\{w \in \reals^n \mid w \ge 0, ~ \ones^Tw = 1\}$.

\paragraph{Convergence.}
When the problem is convex and feasible, the ADMM steps are guaranteed to
converge to a global solution for any penalty parameter $\rho>0$.
When the problem is not convex, the ADMM steps are not guaranteed
to converge to a global solution, or even converge.
However, it has been observed that ADMM is often effective at finding
approximate solutions to nonconvex problems
\cite{diamond2018general,angerisComputationalBoundsPhotonic2019}. 

\paragraph{Efficient computation of steps.}
We note that the first three steps in~\eqref{eq:admm-steps},
which update $f$, $\tilde w$, and $\bar w$,
are independent and can be parallelized, \ie, carried out 
simultaneously.  
Moreover, when $\ell$ or $r$ is separable or block separable, the proximal 
operators can be evaluated on each block independently, leading to 
additional parallelization.

The update for $w^{k+1}$ can be expressed as
the solution to the linear system
\[
\begin{bmatrix}
2I & F^T \\
F & -I
\end{bmatrix}
\begin{bmatrix}
w^{k+1} \\ \eta
\end{bmatrix} = \begin{bmatrix}
F^T(f^{k+1} + y^k) + \tilde w^{k+1} + z^k + \bar w^{k+1} + u^k \\
0 \end{bmatrix}.
\]
The coefficient matrix in this system of linear equations is quasi-definite~\cite{vanderbei1995symmetric},
so it always has an $LDL^T$ factorization with diagonal $D$.
We solve this linear system by performing a sparse $LDL^T$ factorization
and caching the factorization between iterations.
As a result, after the initial factorization, the cost of
each subsequent iteration is that of a back-solve step, which can be 
much smaller than the initial factorization~\cite[App.\ C]{cvxbook}.

There is a large literature on, and many existing libraries for, 
efficiently computing proximal operators; see, \eg,~\cite{parikh2014block,parikh_proximal_code,proximaloperatorsjl}.
We describe here two proximal operators that arise in solving the
representative sample weighting problem, and are less well known.

\paragraph{Proximal operator for KL divergence.}
The proximal operator for KL divergence can arise both in the loss and 
regularizer functions.
The proximal operator of $r(w) = \sum_{i=1}^n w_i \log(w_i / u_i)$
is given by
\[
\mathbf{prox}_{\lambda r}(v)_i = \lambda W\left(u_ie^{v_i/\lambda - 1} / \lambda\right),
\]
where $W$ is the Lambert-$W$ or product log function~\cite{lambert}.

\paragraph{Proximal operator for representative selection.}
In the case where the regularizer $r$ is nonconvex,
its proximal operator is generally difficult to find in practice.
However, one special case where it is in fact easy to evaluate
is in the representative selection problem \eqref{eq:boolean_reg}.
In this case, the proximal operator of $r$
is just the projection of $v$ onto the (nonconvex) set
\[
\{w \in \{0, 1/k\}^n \mid \ones^Tw = 1\}.
\]
This is easily computed by setting the largest $k$ entries of $v$ to $1/k$
(with ties broken arbitrarily)
and the rest to zero~\cite[\S4.2]{diamond2018general}.

\subsection{Implementation}
We have implemented the ADMM algorithm described above
in an easy-to-use Python package \verb|rsw|, which is freely available
online at
\[
\verb|www.github.com/cvxgrp/rsw|.
\]
The current implementation exploits very little of 
the parallelization possible; a future version will exploit much more.

The package exposes a single method, \verb|rsw|, which has the signature
\[
\verb|def rsw(df, funs, losses, reg):|
\]
It performs representative sample weighting on the pandas DataFrame \verb|df|,
using the functions in the list \verb|fun|.
(To use \verb|df| directly as $F$, one would pass in \verb|fun=None|.)
We provide a number of common losses and regularizers, and their associated proximal operators. 
The argument \verb|losses| specifies the losses, applied in block-separable form;
\eg, to express the loss $(f_1 - 1)^2 + I_0(f_2 -2)$, one would pass in
\[
\verb|losses=[rsw.LeastSquaresLoss(1), rsw.EqualityLoss(2)]|.
\]
The argument 1 in \verb|rsw.LeastSquaresLoss(1)| specifies the desired value for $f_1$.
One can also specify the scale of each loss; \eg, for $((1/2)(f_1-1))^2$, one would pass in
the loss
\verb|rsw.LeastSquaresLoss(1, scale=.5)|. 
The argument \verb|reg| specifies the regularizer; \eg, to use entropy regularizer,
one would pass in \verb|reg=rsw.EntropyRegularizer()|; for a Boolean regularizer with 5 nonzeros,
one would pass in \verb|reg=rsw.BooleanRegularizer(5)|.
You can also specify limits on the entries of $w$ as in \eqref{eq:limit}, \eg,
\[
\verb|reg=rsw.EntropyRegularizer(limit = 2)|
\]
constrains $w$ to not be up- or down-weighted from uniform by a factor of $2$,
or $1/(2n) \ones \leq w \leq 2/n\ones$.
At this point, the software package does not support making the loss or regularizer
a sum of multiple losses or regularizers.
We note however that this can easily be accomodated by adding 
additional consensus variables,
and we plan to implement this in a future version of the software.

\paragraph{Packages.}
We use the Python packages \verb|numpy|~\cite{van2011numpy} for dense linear algebra,
\verb|scipy|~\cite{jones2001scipy} for sparse linear algebra,
\verb|pandas|~\cite{mckinney-proc-scipy-2010} for data manipulation,
and \verb|qdldl|~\cite{stellato2020osqp,qdldl} for the sparse quasi-definite $LDL^T$ factorization.

\section{Numerical experiments}
\label{s-experiments}
We present here an application of the methods
described above to a large dataset from the US Center for Disease Control (CDC).
We chose examples that are a bit more complex that might be needed,
to demonstrate the method's flexibility and scalability.
We note that, in practice, the functions and losses used would likely 
be much simpler, although they need not be.
All experiments were conducted on a single core of an Intel i7-8700K CPU
with 64GB of memory, which, at the time of writing, was a standard
consumer desktop CPU.

\paragraph{CDC BRFSS dataset.} Each year, the US Center for Disease Control (CDC)
conducts health-related telephone surveys of hundreds of thousands
of US citizens.
(This is the world's largest telephone survey.)
The data is collected as part of the Behavioral Risk Factor Surveillance System (BRFSS),
and is freely available online~\cite{brfss}.
We consider the 2018 version of this dataset,
which contains 437436 responses to 275 survey questions.
The data includes a variety of demographic and health-related attributes~\cite{llcp_codebook}.
We consider the following columns in the dataset:
\begin{itemize}
	\item \emph{State or territory}. There are 53 possible values; \eg, NM (New Mexico) or PR (Puerto Rico).
	\item \emph{Age group}. The 6 possible values are 18-24, 25-34, 35-44, 45-54, 55-65, and 65+.
	\item \emph{Sex}. The 2 possible values are male and female.
	\item \emph{Education level.} The 4 possible values are elementary, high school, some college, and college.
	\item \emph{Income.} The 8 possible values are 0-10k, 10-15k, 15-20k, 20-25k, 25-35k, 35-50k, 50-75k, and 75k+.
	\item \emph{Reported general health.} The 5 possible values are excellent, very good, good, fair, and poor.
\end{itemize}

\paragraph{Functions.}
As our first group of functions, we consider the $\{0,1\}$ indicator functions
of each possibility of state and age group, of which there are 318.
(The expected values of these functions are the probabilities of each
state/age group).
Our next two functions are the indicator function of the sex being male
or female.
Our next group of functions are the indicator functions of each
possibility of education level and income, of which there are 32.
The final group of functions are the indicator functions of each
possibility of reported general health, of which there are 5.
In total, we have $m=357$ functions.
The resulting matrix $F$ has over 150 million entries and around 2\%
of its values are missing.
It is sparse, with a density of 3.5\%.

Our choice of these particular 357 functions is relatively arbitrary
(although they are not unreasonable).

\paragraph{Desired expected values.}
In all experiments, we take the averages across the entire dataset
as the desired expected values, or
\[
f^\mathrm{des} = (1/n)F\ones.
\]
We ignore missing entries in the computation.
The desired expected values are given
in tables that can be found in appendix \ref{sec:desired_tables}.

\paragraph{Constructing a skewed subsample.}
We generate a smaller dataset composed of 10000 samples
by generating a skewed sample from the dataset.
Denote the $i$th column of $F$ by $x_i\in\reals^m$.
As described above, $x_i$ contains a one-hot encoding of various
things (state and age, sex, education and income, and health).
We generate a skewed sampling distribution from the dataset by first
sampling $c\sim\mathcal N(0, (1/4)I)$, and then letting
\[
\pi_i = \exp(c^Tx_i) / \sum_{j=1}^n \exp(c^Tx_j), \quad i=1,\ldots,n.
\]
(By construction, $\pi \geq 0$ and $\ones^T\pi=1$.)
Suppose entry $j$ of the samples is the indicator function for female samples.
If, \eg, $c_j = -0.1$, then we are less likely to sample females, since
whenever $(x_i)_j=1$, this subtracts 0.1 from the log probability of
sample $i$, giving the female samples a smaller chance of being sampled.
On the other hand, if $c_j=0.1$, then samples that are female
get 0.1 added to their log probabilities.
Next we take 10000 samples without replacement from the columns of $F$
using the distribution $\pi$.
This sample is highly skewed; for example, 41.2\% of samples are female,
whereas in the original dataset 54.8\% samples are female.
Therefore, we need to do sample weighting to make
the sample representative.

\subsection{Maximum entropy weighting}
\label{sec:max_entr_weighting_experiment}
\begin{figure}
\centering
\includegraphics[width=.5\textwidth]{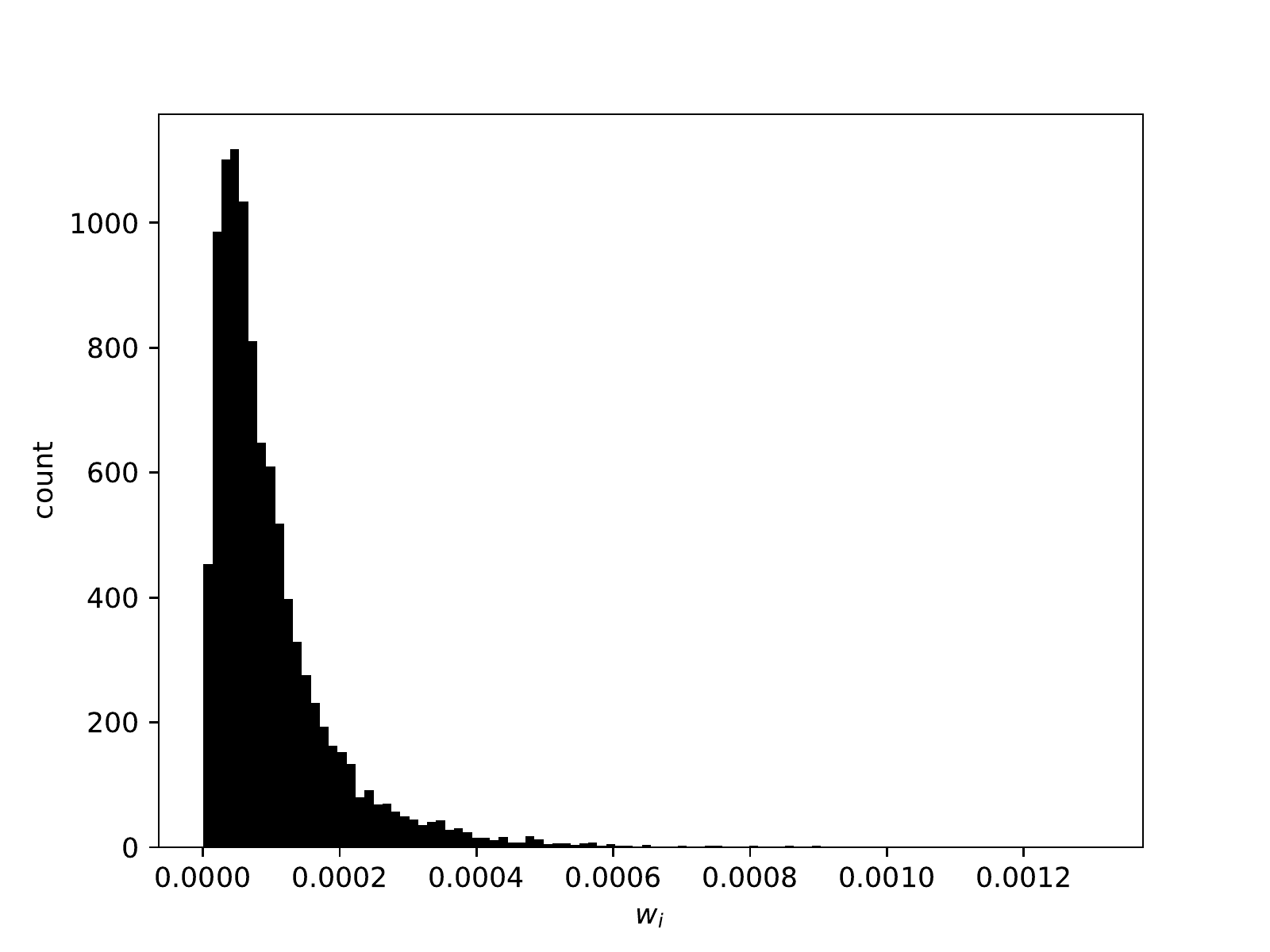}
\caption{Histogram of maximum entropy weights.}
\label{fig:hist_w}
\end{figure}

We first find the maximum entropy weighting of these 10000
samples such that the induced expected values match
the desired expected values, \ie, we solve the problem
described in \S\ref{sec:max-entropy}.
We solved the problem using our ADMM implementation,
which took around 21 seconds.
(Using a generic cone program solver like SCS, the default
solver for CVXPY, took 19 minutes.)
The resulting entropy of $w$ was 8.59, whereas the entropy of the uniform
distribution on $\{1,\ldots,10000\}$ is 9.21.
The histogram of $w$ is shown in figure \ref{fig:hist_w},
and appears to have a long right tail, meaning that some samples
are considerably up-weighted.

\begin{figure}
\centering
\includegraphics[width=.5\textwidth]{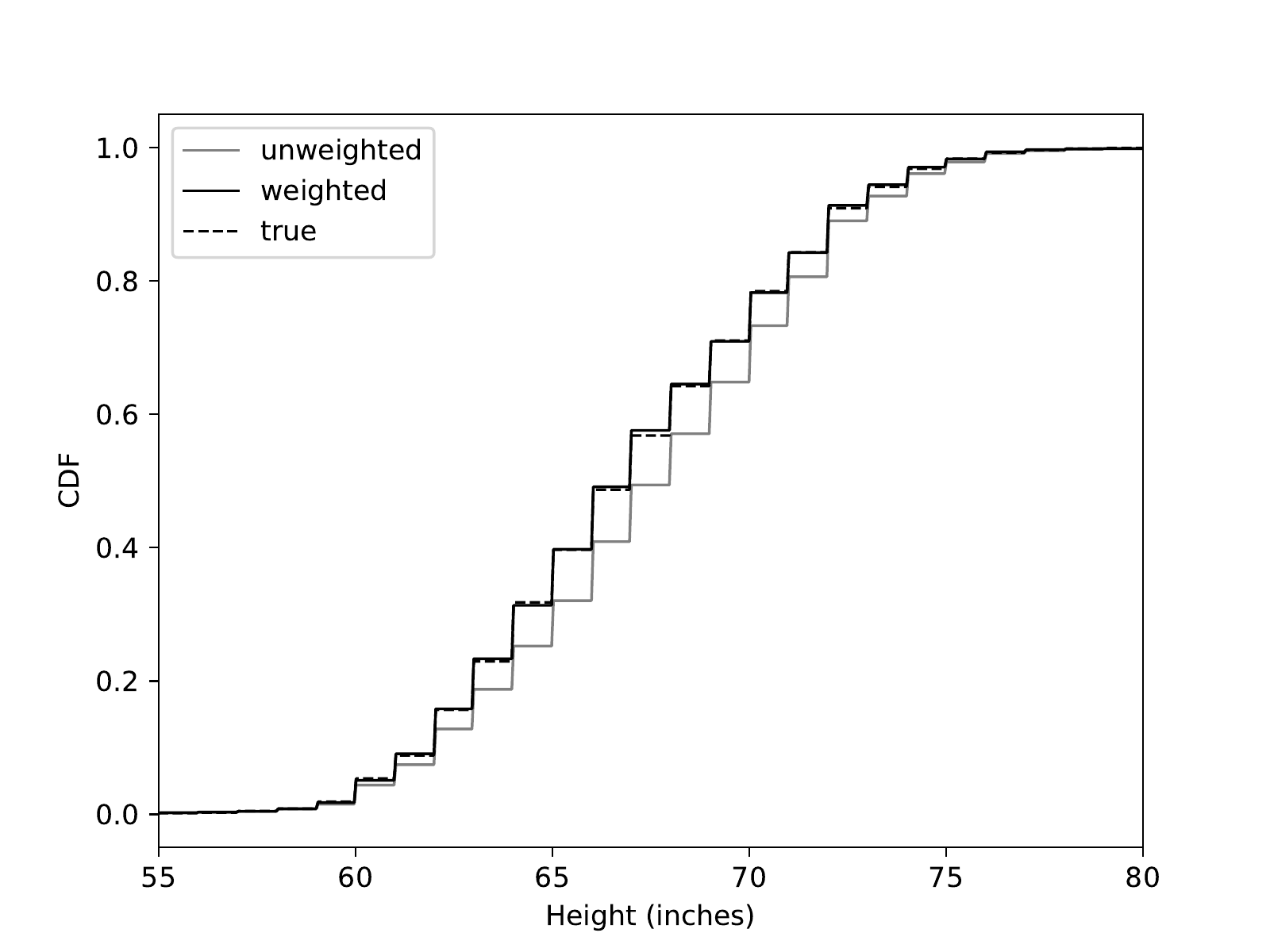}
\caption{CDF of height for unweighted, weighted, and true distributions.}
\label{fig:height_cdf}
\end{figure}

\begin{figure}
\centering
\includegraphics[width=.5\textwidth]{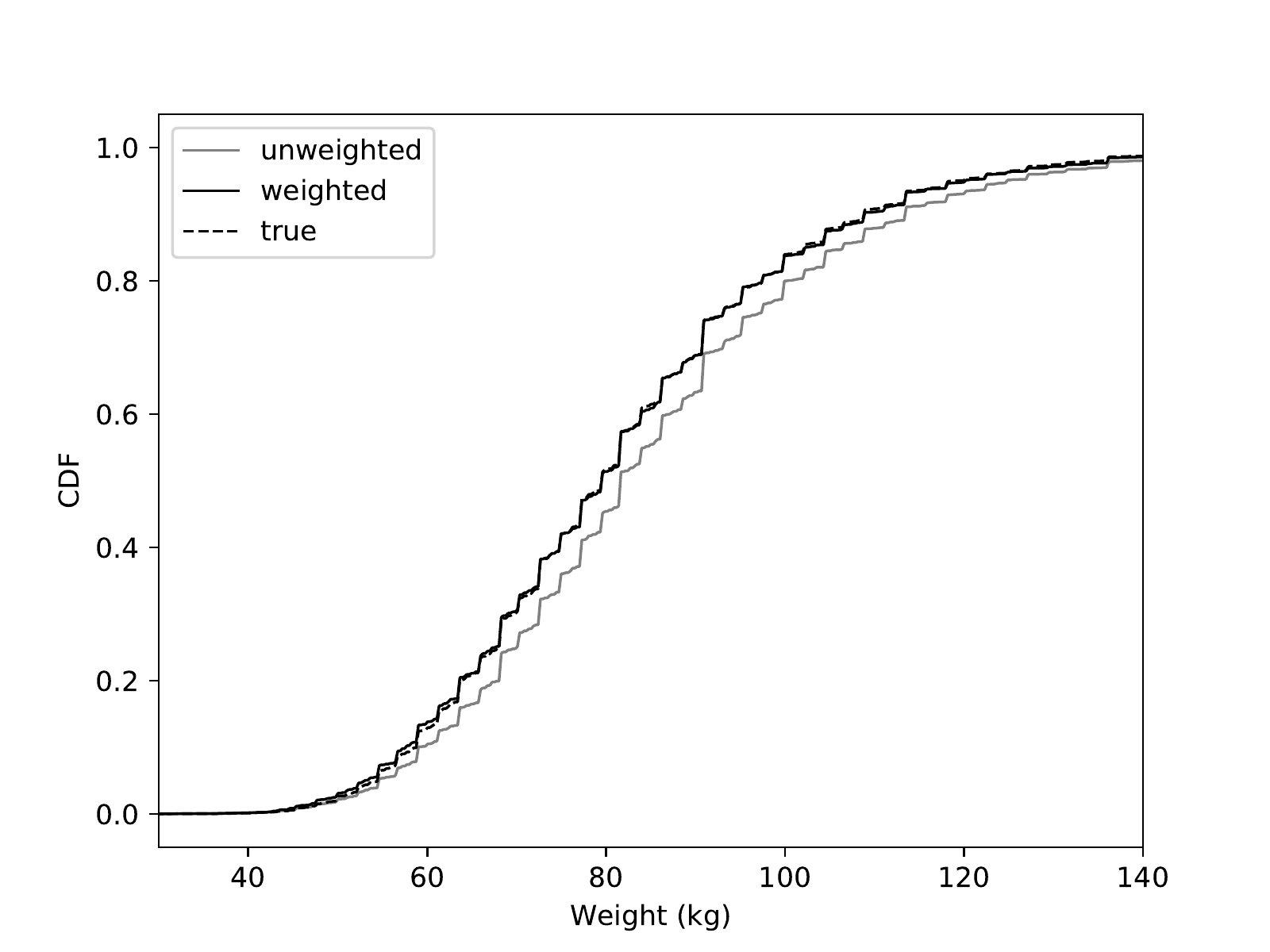}
\caption{CDF of weight for unweighted, weighted, and true distributions.}
\label{fig:weight_cdf}
\end{figure}

To illustrate a use for these weights, we compare the distributions
of other columns not included in the procedure to the true distribution
across the entire dataset.
We did this for the height and weight columns,
and in figures \ref{fig:height_cdf} and \ref{fig:weight_cdf}
we plot the cumulative distribution function (CDF) of height and weight in the true
distribution, and the unweighted and weighted distribution over
the subsample.
The weighted CDF appears to match the true CDF much
better than the unweighted CDF.
We computed the Kolmogorov–Smirnov (K-S) test statistic~\cite{kolmogorov1933sulla},
\ie, the maximum absolute difference between the CDFs,
between the unweighted and true CDF and between the weighted and true CDF.
For the height attribute, the K-S statistic was 0.078 for the unweighted
distribution, and 0.008 for the weighted distribution, which is almost
a factor of 10 better.
For the weight attribute, the K-S statistic was 0.064 for the unweighted
distribution, and 0.009 for the weighted distribution, a factor of 
more than 7 better.

We remind the reader that our focus here is not on using the weights,
but only on how to compute them.  We give these results only to show
that weighting has indeed improved our estimate of 
the CDF of unrelated functions of the samples.  We also note that 
a similar improvement in estimating the distribution of height and
weight can be obtained with a far simpler representative
sample weighting problem, for example one that takes into account
only sex and age group.

\subsection{Representative selection}
\begin{figure}
\centering
\includegraphics[width=.5\textwidth]{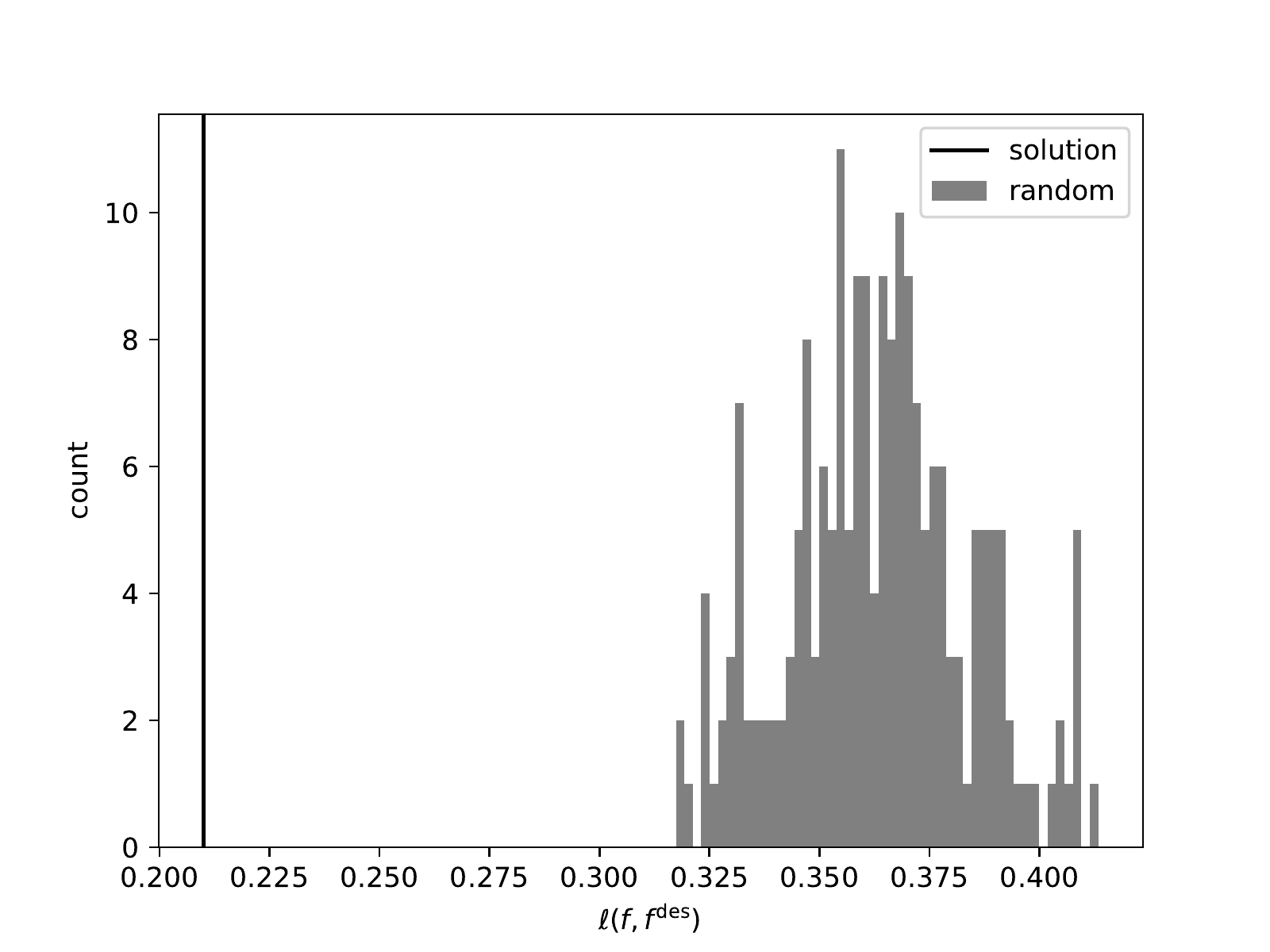}
\caption{Histogram of losses for random subsamples and our selection (as a vertical line).}
\label{fig:boolean}
\end{figure}

\begin{figure}
\centering
\includegraphics[width=.5\textwidth]{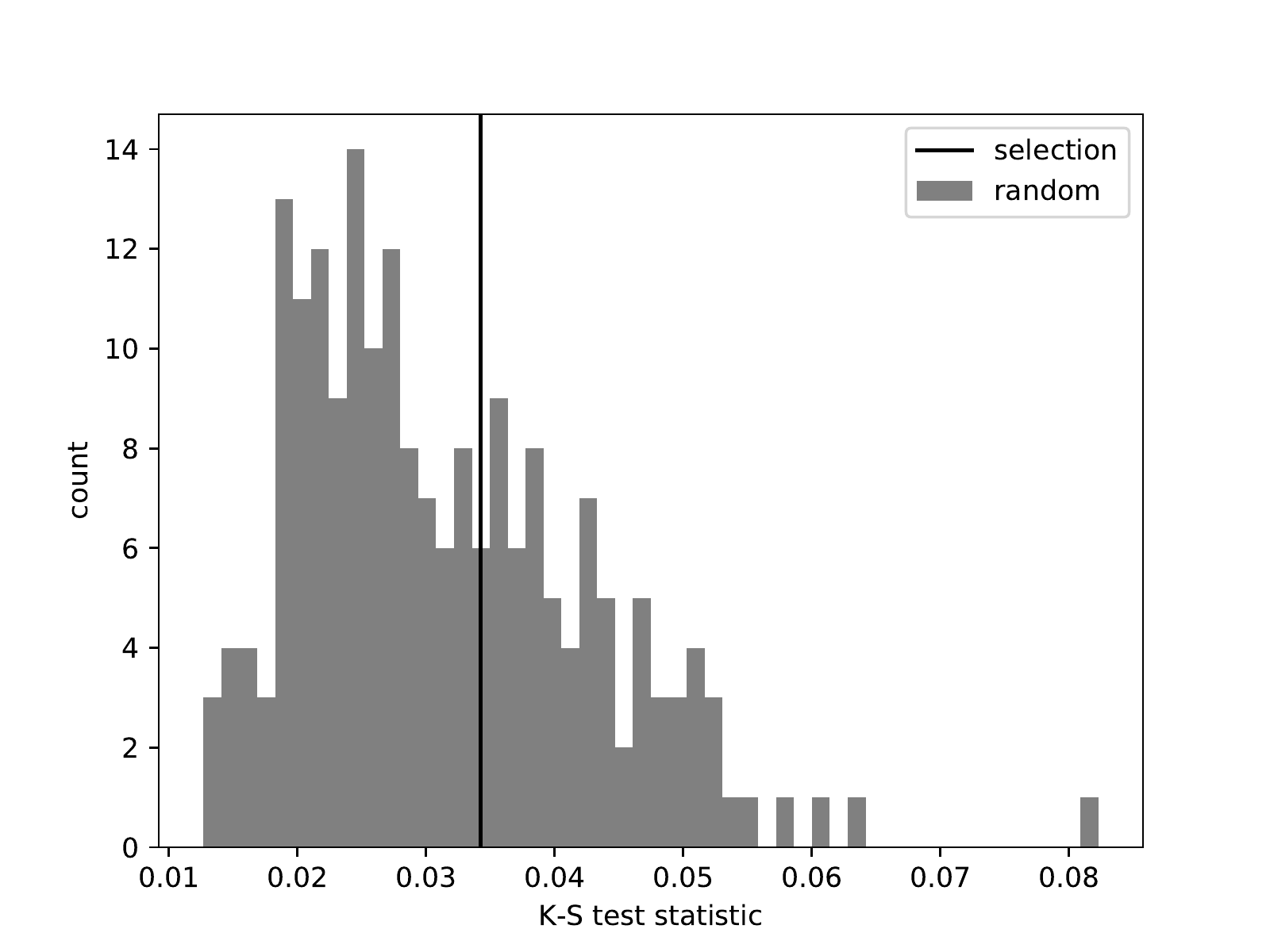}
\caption{Histogram of KS-test statistic for height.}
\label{fig:ks_boolean_height}
\end{figure}

\begin{figure}
\centering
\includegraphics[width=.5\textwidth]{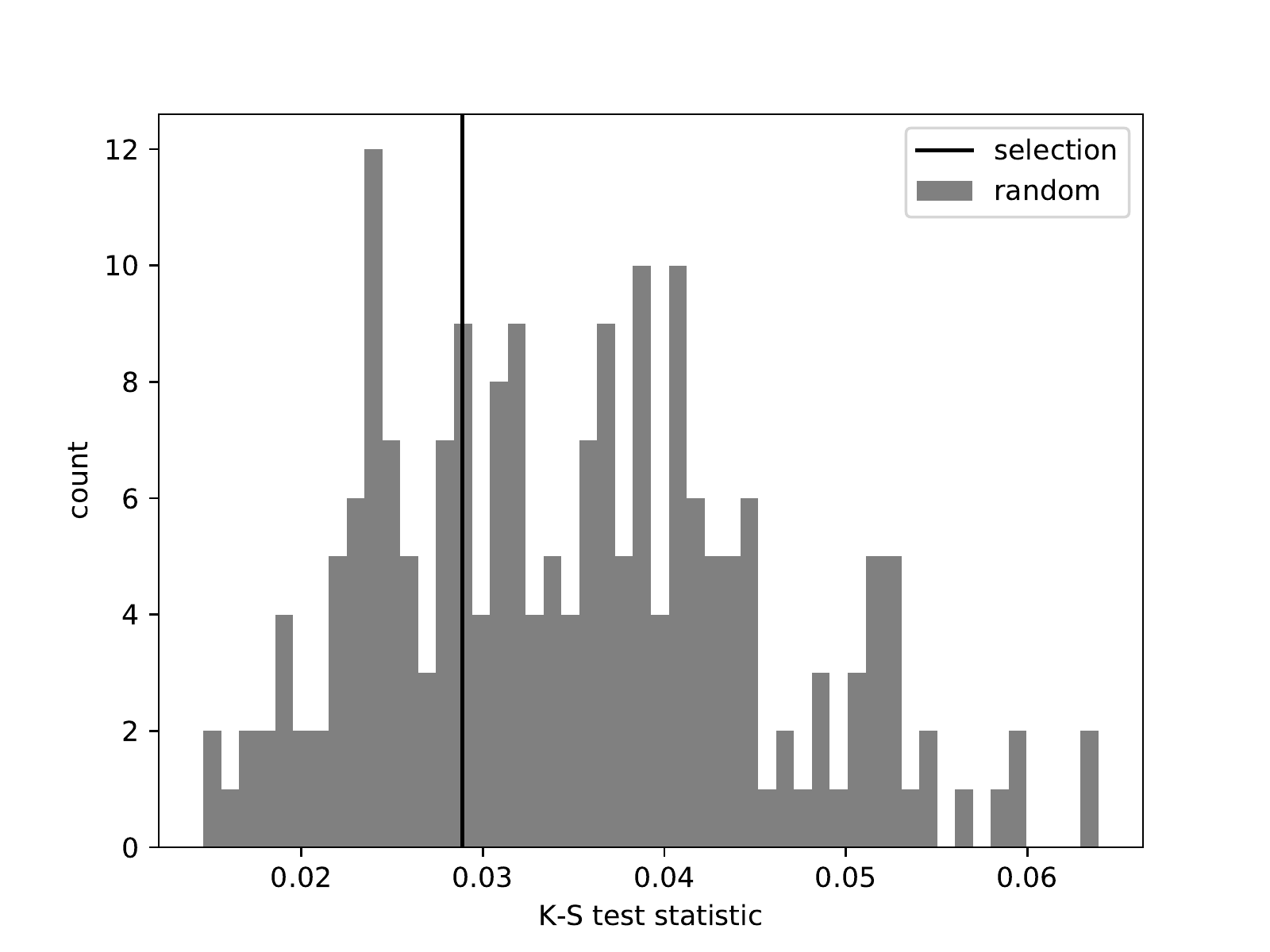}
\caption{Histogram of KS-test statistic for weight.}
\label{fig:ks_boolean}
\end{figure}

Next we consider the representative selection problem,
described in \S\ref{sec:representative_selection},
selecting $k=500$ samples with uniform weights
from the $n=10000$ samples.
Since we can no longer match $f^\mathrm{des}$ exactly,
as the loss we use the sum of the KL divergence for each of state-age,
sex, education-income, and general health.
We approximately solved the problem using our ADMM implementation,
initialized with a uniform distribution,
which took under 12 seconds. The final loss of our weights was 0.21.
We compared the loss to the loss of 200 random subsamples, which were each generated
by sampling without replacement with the maximum entropy weights;
the histogram of these objective values are displayed in figure \ref{fig:boolean}.
Our solution has substantially lower loss than even the best random subsample.
We also computed the K-S statistic for the CDF of the height and weight columns
between the 200 random samples and the true distribution, and between
our selection and the true distribution.
The histogram of the test statistic, as well as our test statistic,
for the height and weight CDFs are
displayed in figure \ref{fig:ks_boolean_height} and \ref{fig:ks_boolean};
we had a lower test statistic 36.5\% of the time for height
and 68.5\% of the time for weight.
So our selection is less representative in terms of height
than if we had sampled using the maximum entropy weights,
but more representative in terms of weight.
(Both of these methods are far superior to selecting $k$ samples
uniformly at random from the 10000.)

\section{Extensions and variations}
\label{s-ext-var}

In this section we discuss extensions and variations.

\paragraph{Missing entries in some data.}
Here we describe what to do when some entries of $F$ are missing,
\ie, $F_{ij}\in\reals \cup \{?\}$, where $?$ denotes a missing value.
The simplest and often most effective thing to do when there are missing
values is to replace them with the desired expected value for that function.
That is, suppose one of our functions is the indicator function of a sample being female.
If, for one sample, we did not know their sex, we would replace that missing
value in $F$ with the desired proportion of females.
Assuming the loss is separable and homogeneous, which all aforementioned losses are,
filling in missing entries this way has the same effect as
ignoring the missing entries in the computation of $f-f^\mathrm{des}$,
and scaling the loss by some multiple of one minus the sum of the weights
for samples with missing entries of that feature.
In our implementation, we fill in missing entries this way.

\paragraph{Conditional probabilities.}
Suppose $F_1$ is the $\{0,1\}$ indicator function
of some set $A \cap B$, where $A,B\subset\mathcal X$,
so $f_1$ is the probability that $x\in A \cap B$,
and $F_2$ is the $\{0,1\}$ indicator function
of $B$,
so $f_2$ is the probability that $x\in B$.
Then the quantity $f_1/f_2$
is equal to the (conditional) probability that $x\in A$ given that
$x\in B$.
For example, if $A$ contains the samples that are over the age of
65, and $B$ contains the samples that are female,
then $f_1/f_2$ is the probability that the sample is over the age of
65 given it is female.
If we want the induced probability $f_1/f_2$ to be close
to some value $f^\mathrm{des}$, we can use the loss
\[
l(f, f^\mathrm{des}) = l_0(f_1 - f_2 f^\mathrm{des}),
\]
where $l_0$ is some convex function.
The function $l$ is convex because $f_1 - f_2 f^\mathrm{des}$
is affine in $w$.

\paragraph{CDFs.}
Suppose $x\in\reals$.
We can make $f$ the CDF evaluated at $\alpha_1,\ldots,\alpha_m\in\reals$
by letting $F_i$ be the \{0,1\} indicator function of the set
$\{x \mid x \leq \alpha_i\}$, $i=1,\ldots,m$.
Then $f^\mathrm{des}$ would be the desired CDF.
A common loss function
is the K-S test statistic, or
\[
l(f,f^\mathrm{des}) = \|f-f^\mathrm{des}\|_\infty,
\]
which is convex.

\section*{Acknowledgements}
The authors would like to thank Trevor Hastie,
Timothy Preston, Jeffrey Barratt, and Giana Teresi for discussions about the ideas
described in this paper.
Shane Barratt is supported by the National Science Foundation Graduate Research Fellowship
under Grant No. DGE-1656518.

\clearpage
\bibliography{citations}

\clearpage
\appendix

\section{Iterative proportional fitting}\label{app:coordinate-ascent}
The connection between iterative proportional fitting, initially proposed
by~\cite{deming1940least} and the maximum entropy weighting problem has long
been known and has been explored by many authors~\cite{teh2003improving,
fu2019cvxr, she2019iterative, wittenbergIntroductionMaximumEntropy2010,
bishopDiscreteMultivariateAnalysis2007}. We provide a similar presentation
to~\cite[\S2.1]{she2019iterative}, though we show that the iterative
proportional fitting algorithm that is commonly implemented is actually a block
coordinate descent algorithm on the dual variables, rather than a direct
coordinate descent algorithm. Writing this update in terms of the primal
variables gives exactly the usual iterative proportional fitting update over
the marginal distribution of each property.

\paragraph{Maximum entropy problem.}
In particular, we will analyze the application of block coordinate descent on
the dual of the problem
\begin{equation}\label{eq:max-ent-problem}
\begin{array}{ll}
	\mbox{minimize} & \sum_{i=1}^n w_i \log w_i\\
	\mbox{subject to} & Fw = f^\mathrm{des}\\
	& \ones^Tw = 1, ~~ w \ge 0,
\end{array}
\end{equation}
with variable $w \in \reals^n$, where the problem data matrix is Boolean, \ie,
$F \in \{0, 1\}^{m \times n}$.  This is just the maximum entropy weighting
problem given in~\S\ref{sec:max-entropy}, but in the specific case where $F$ is
a matrix with Boolean entries.

\paragraph{Selector matrices.} We will assume that we have several possible
categories $k=1, \dots, N$ which the user has stratified over, and we will
define selector matrices $S_k \in \{0,1\}^{p_k \times m}$ which `pick out' the
rows of $F$ containing the properties for property $k$. For example, if the
first three rows of $F$ specify the data entries corresponding to the first
property, then $S_1$ would be a matrix such that
\[
S_1F = F_{1:3,1:n},
\]
and each column of $S_1F$ is a unit vector; \ie, a vector whose entries are all
zeros except at a single entry, where it is one. This is the same as saying
that, for some property $k$, each data point is allowed to be in exactly one of
the $p_k$ possible classes. Additionally, since this should be a proper
probability distribution, we will also require that $\ones^TS_k f^\mathrm{des}
= 1$, \ie, the desired marginal distribution for property $k$ should itself sum
to 1.

\paragraph{Dual problem.}
To show that iterative proportional fitting is equivalent to block coordinate
ascent, we first formulate the dual problem~\cite[Ch.~5]{cvxbook}.
The Lagrangian of~\eqref{eq:max-ent-problem} is
\[
\mathcal{L}(w, \nu, \lambda) = \sum_{i=1}^n w_i \log w_i + \nu^T(Fw - f^\mathrm{des}) + \lambda (\ones^Tw - 1),
\]
where $\nu \in \reals^n$ is the dual variable for the first constraint and
$\lambda \in \reals$ is the dual variable for the normalization constraint.
Note that we do not need to include the nonnegativity constraint on $w_i$,
since the domain of $w_i \log w_i$ is $w_i \ge 0$.

The dual function~\cite[\S5.1.2]{cvxbook} is given by
\begin{equation}\label{eq:dual-function}
g(\nu, \lambda) = \inf_{w} \mathcal{L}(w, \nu, \lambda),
\end{equation}
which is easily computed using the Fenchel conjugate of the negative
entropy~\cite[\S3.3.1]{cvxbook}:
\begin{equation}\label{eq:dual-function-full}
g(\nu, \lambda) = - \ones^T\exp(-(1 + \lambda) \ones - F^T\nu) - \nu^Tf^\mathrm{des} - \lambda,
\end{equation}
where $\exp$ of a vector is interpreted componentwise. Note that the optimal weights $w^\star$
are exactly those given by
\begin{equation}\label{eq:w-optimal}
w^\star = \exp( -(1 + \lambda) \ones - F^T\nu).
\end{equation}

\paragraph{Strong duality.} Because of strong duality, the maximum of the dual
function~\eqref{eq:dual-function} has the same value as the optimal value of
the original problem~\eqref{eq:max-ent-problem}~\cite[\S5.2.3]{cvxbook}.
Because of this, it suffices to find an optimal pair of dual variables,
$\lambda$ and $\nu$, which can then be used to find an optimal $w^\star$,
via~\eqref{eq:w-optimal}.

To do this, first partially maximize $g$ with respect to $\lambda$, \ie,
\[
g^p(\nu) = \sup_{\lambda} g(\nu, \lambda).
\]
We can find the minimum by differentiating~\eqref{eq:dual-function-full} with respect to $\lambda$ and
setting the result to zero. This gives
\[
1+\lambda^\star = \log\left(\ones^T\exp(-F^T\nu)\right),
\]
while
\[
g^p(\nu) = -\log(\ones^T\exp(-F^T\nu)) - \nu^Tf^\mathrm{des}.
\]
This also implies that, after using the optimal $\lambda^\star$ in~\eqref{eq:w-optimal},
\begin{equation}\label{eq:w-optimal-normalized}
w^\star = \frac{\exp(- F^T\nu)}{\ones^T\exp(- F^T\nu)}.
\end{equation}

\paragraph{Block coordinate ascent.} In order to maximize the dual function
$g^p$, we will use the simple method of block coordinate ascent with respect to
the dual variables corresponding to the constraints of each of the possible $k$
categories. Equivalently, we will consider updates of the form
\[
\nu^{t + 1} = \nu^{t} + S_{t}^T\xi^t,\quad 
t =1, \ldots, T,
\]
where $\nu^t$ is the dual variable at iteration $t$, while $\xi^t \in
\reals^{p_t}$ is the optimization variable we consider at iteration $t$.
To simplify notation, we have used $S_t$ to refer to the selector matrix at
iteration $t$, if $t \le N$, and otherwise set $S_t = S_{(t-1 \mod
N) + 1}$; \ie, we choose the selector matrices in a round robin fashion. The
updates result in an ascent algorithm, which is guaranteed to converge to the
global optimum since $g^p$ is a smooth concave
function~\cite{tsengConvergenceBlockCoordinate2001}.

\paragraph{Block coordinate update.} In order to apply the update rule to
$g^p(\nu)$, we first work out the optimal steps defined as
\[
\xi^{t} = \argmax_{\xi} ~g^p(\nu^t + S^T_t \xi).
\]
To do this, set the gradient of $g_p$ to zero,
\[
\nabla_\xi ~g^p(\nu^t + S^T_t \xi) = 0,
\]
which implies that
\begin{equation}\label{eq:dual-block-opt}
\frac{\sum_{i=1}^n (S_t f_i) \exp(-f_i^T\nu^t - f_i^TS_t^T\xi)}{\sum_{i=1}^n \exp(-f_i^T\nu^t - f_i^T S^T_t \xi)} = Sf^\mathrm{des},
\end{equation}
where $f_i$ is the $i$th column of $F$ and the division is understood to be elementwise.

To simplify this expression, note that, for any unit basis vector $x \in
\reals^m$ (\ie, $x_i = 1$ for some $i$ and 0, otherwise), we have the simple
equality,
\[
x\exp(x^T\xi) = x\circ \exp(\xi),
\]
where $\circ$ indicates the elementwise product of two vectors. Using this
result with $x = S_t f_i$ on each term of the numerator from the left hand
side of~\eqref{eq:dual-block-opt} gives
\[
\sum_{i=1}^n (S_t f_i) \exp(-f_i^T\nu^t - f_i^TS_t^T\xi) = \exp(-\xi) \circ \left(\sum_{i=1}^n \exp(-f_i^T\nu^t)S_t f_i\right) = \exp(-\xi) \circ y,
\]
where we have defined $y = \sum_{i=1}^m \exp(-f_i^T\nu^t)Sf_i$ for
notational convenience. We can then rewrite~\eqref{eq:dual-block-opt} in terms
of $y$ by multiplying the denominator on both sides of the expression:
\[
\exp(-\xi) \circ y = (\exp(-\xi)^Ty)S_t f^\mathrm{des},
\]
which implies that
\[
\frac{y \circ \exp(-\xi)}{\ones^T(y \circ \exp(-\xi))} = S_t f^\mathrm{des}.
\]
Since $\ones^TS_t f^\mathrm{des} = 1$, then
\[
y \circ \exp(-\xi) = S_t f^\mathrm{des},
\]
or, after solving for $\xi$,
\[
\xi = -\log\left(\diag(y)^{-1}S_t f^\mathrm{des}\right),
\]
where the logarithm is taken elementwise. The resulting block coordinate ascent 
update can be written as
\begin{equation}\label{eq:nu-update}
\nu^{t+1} = \nu^t - S_t^T \log\left(\frac{S_t f^\mathrm{des}}{\sum_{i=1}^n \exp(-f_i^T\nu^t)S_t f_i}\right),
\end{equation}
where the logarithm and division are performed elementwise. This update can be
interpreted as changing $\nu$ in the entries corresponding to the constraints
given by property $t$ by adding the log difference between the desired
distribution and the (unnormalized) marginal distribution for this property
suggested by the previous update. This follows
from~\eqref{eq:w-optimal-normalized}, which implies $w_i^t \propto
\exp(-f_i^T\nu^t)$ for each $i=1, \dots, n$, where $w^t$ is the
distribution suggested by $\nu^t$ at iteration $t$.

\paragraph{Resulting update over $w$.}  We can rewrite the update for the dual
variables $\nu$ as a multiplicative update for the primal variable $w$, which
is exactly the update given by the iterative proportional fitting algorithm.
More specifically, from~\eqref{eq:w-optimal-normalized},
\[
w^{t+1}_i = \frac{\exp(-f_i^T\nu^{t+1})}{\sum_{i=1}^n \exp(-f_i^T\nu^{t+1})}.
\]
For notational convenience, we will write $x_{t i}= S_t f_i$, which is a
unit vector denoting the category to which data point $i$ belongs to, for
property $t$. Plugging update~\eqref{eq:nu-update} gives, after 
some algebra,
\[
\exp(-f_i^T\nu^{t+1}) = \exp(-f_i^T\nu^t) \frac{\exp(x_{t
i}^T\log(S_t f^\mathrm{des}))}{\exp\left(x_{t i}^T\log(\sum_{j=1}^n
\exp(-f_j^T\nu^t)x_{t j})\right)}.
\]
Since $x_{t i}$ is a unit vector, then $ \exp(x_{t i}^T \log(y)) = x_{t i}^Ty$ for any vector $y > 0$, so
\[
\exp(-f_i^T\nu^{t+1}) = \exp(-f_i^T\nu^t) \frac{x_{t i}^TS_t
f^\mathrm{des}}{\sum_{j=1}^n \exp(-f_j^T\nu^t)x_{t i}^Tx_{t j}}.
\]
Finally, using~\eqref{eq:w-optimal-normalized} with $\nu^t$ gives
\[
w_i^{t+1} = w_i^t\frac{x_{t i}^TS_t f^\mathrm{des}}{\sum_{j=1}^n w^t_j (x_{t i}^Tx_{t j})},
\]
which is exactly the multiplicative update of the iterative proportional
fitting algorithm, performed for property $t$.

\clearpage
\section{Expected values of BRFSS data}
\label{sec:desired_tables}

\begin{table}[h]
\caption{Desired sex values in percentages.}
\centering
\begin{tabular}{ll}
\toprule
Male & Female \\
\midrule
45.2 & 54.8 \\
\bottomrule
\end{tabular}
\end{table}

\begin{table}[h]
\caption{Desired education and income values in percentages.}
\centering
\begin{tabular}{lllllllll}
\toprule
Education & 0-10k & 10-15k & 15-20k & 20-25k & 25-35k & 35-50k & 50-75k & 70k+ \\
\midrule
Elementary & 0.455 & 0.355 & 0.432 & 0.344 & 0.266 & 0.168 & 0.084 & 0.096 \\
High School & 1.877 & 2.059 & 3.064 & 3.702 & 3.941 & 4.313 & 3.876 & 4.730 \\
Some College & 1.172 & 1.440 & 2.098 & 2.861 & 3.602 & 4.752 & 5.198 & 7.992 \\
College & 0.552 & 0.666 & 1.024 & 1.653 & 2.547 & 4.691 & 7.415 & 22.576 \\
\bottomrule
\end{tabular}
\end{table}

\begin{table}[h]
\caption{Desired reported health values in percentages.}
\centering
\begin{tabular}{llllll}
\toprule
Excellent & Very Good & Good & Fair & Poor \\
\midrule
16.48 & 32.6 & 31.7 & 13.93 & 5.3 \\
\bottomrule
\end{tabular}
\end{table}

\begin{table}
\caption{Desired state and age values in percentages. Continued on next page.}
\centering
\begin{tabular}{lllllll}
\toprule
State & 18-24 & 25-34 & 35-44 & 45-54 & 55-65 & 65+ \\
\midrule
AK & 0.040 & 0.069 & 0.083 & 0.091 & 0.147 & 0.200 \\
AL & 0.070 & 0.139 & 0.176 & 0.239 & 0.332 & 0.553 \\
AR & 0.039 & 0.085 & 0.099 & 0.152 & 0.240 & 0.610 \\
AZ & 0.095 & 0.184 & 0.195 & 0.259 & 0.374 & 0.741 \\
CA & 0.260 & 0.453 & 0.458 & 0.443 & 0.486 & 0.625 \\
CO & 0.127 & 0.243 & 0.262 & 0.310 & 0.410 & 0.683 \\
CT & 0.102 & 0.200 & 0.241 & 0.435 & 0.558 & 0.913 \\
DC & 0.042 & 0.107 & 0.128 & 0.166 & 0.188 & 0.374 \\
DE & 0.072 & 0.120 & 0.144 & 0.185 & 0.253 & 0.422 \\
FL & 0.201 & 0.367 & 0.369 & 0.509 & 0.657 & 1.382 \\
GA & 0.157 & 0.257 & 0.290 & 0.346 & 0.419 & 0.658 \\
GU & 0.035 & 0.051 & 0.055 & 0.077 & 0.076 & 0.081 \\
HI & 0.119 & 0.212 & 0.241 & 0.291 & 0.361 & 0.582 \\
IA & 0.148 & 0.228 & 0.280 & 0.330 & 0.426 & 0.673 \\
ID & 0.062 & 0.096 & 0.098 & 0.110 & 0.155 & 0.326 \\
IL & 0.097 & 0.163 & 0.171 & 0.203 & 0.235 & 0.344 \\
IN & 0.084 & 0.145 & 0.168 & 0.274 & 0.370 & 0.698 \\
KS & 0.151 & 0.249 & 0.275 & 0.371 & 0.503 & 0.924 \\
\bottomrule
\end{tabular}
\end{table}

\begin{table}
\caption{Desired state and age values in percentages.}
\centering
\begin{tabular}{lllllll}
\toprule
State & 18-24 & 25-34 & 35-44 & 45-54 & 55-65 & 65+ \\
\midrule
KY & 0.129 & 0.192 & 0.218 & 0.332 & 0.436 & 0.647 \\
LA & 0.083 & 0.150 & 0.157 & 0.204 & 0.256 & 0.322 \\
MA & 0.115 & 0.184 & 0.176 & 0.249 & 0.306 & 0.495 \\
MD & 0.147 & 0.294 & 0.386 & 0.646 & 0.884 & 1.655 \\
ME & 0.073 & 0.165 & 0.206 & 0.342 & 0.558 & 1.164 \\
MI & 0.187 & 0.280 & 0.284 & 0.374 & 0.488 & 0.746 \\
MN & 0.277 & 0.459 & 0.543 & 0.639 & 0.818 & 1.148 \\
MO & 0.064 & 0.128 & 0.137 & 0.185 & 0.283 & 0.627 \\
MS & 0.078 & 0.135 & 0.170 & 0.221 & 0.296 & 0.437 \\
MT & 0.066 & 0.113 & 0.140 & 0.158 & 0.261 & 0.449 \\
NC & 0.086 & 0.144 & 0.146 & 0.186 & 0.192 & 0.327 \\
ND & 0.058 & 0.108 & 0.123 & 0.166 & 0.289 & 0.541 \\
NE & 0.211 & 0.353 & 0.417 & 0.450 & 0.669 & 1.245 \\
NH & 0.039 & 0.081 & 0.119 & 0.202 & 0.319 & 0.551 \\
NJ & 0.048 & 0.095 & 0.106 & 0.115 & 0.139 & 0.203 \\
NM & 0.092 & 0.167 & 0.185 & 0.222 & 0.337 & 0.532 \\
NV & 0.058 & 0.095 & 0.092 & 0.115 & 0.135 & 0.242 \\
NY & 0.506 & 0.945 & 1.025 & 1.384 & 1.744 & 2.572 \\
OH & 0.151 & 0.263 & 0.290 & 0.438 & 0.668 & 1.108 \\
OK & 0.063 & 0.126 & 0.142 & 0.163 & 0.246 & 0.461 \\
OR & 0.097 & 0.179 & 0.192 & 0.233 & 0.250 & 0.408 \\
PA & 0.110 & 0.190 & 0.190 & 0.237 & 0.282 & 0.413 \\
PR & 0.080 & 0.151 & 0.165 & 0.198 & 0.213 & 0.300 \\
RI & 0.058 & 0.112 & 0.144 & 0.208 & 0.294 & 0.466 \\
SC & 0.102 & 0.181 & 0.239 & 0.332 & 0.503 & 1.122 \\
SD & 0.084 & 0.160 & 0.173 & 0.234 & 0.345 & 0.633 \\
TN & 0.069 & 0.139 & 0.156 & 0.191 & 0.248 & 0.377 \\
TX & 0.168 & 0.332 & 0.327 & 0.373 & 0.463 & 0.902 \\
UT & 0.241 & 0.376 & 0.434 & 0.382 & 0.389 & 0.591 \\
VA & 0.133 & 0.251 & 0.285 & 0.395 & 0.511 & 0.783 \\
VT & 0.052 & 0.114 & 0.162 & 0.247 & 0.363 & 0.557 \\
WA & 0.155 & 0.329 & 0.364 & 0.415 & 0.590 & 1.146 \\
WI & 0.069 & 0.119 & 0.133 & 0.184 & 0.253 & 0.384 \\
WV & 0.050 & 0.091 & 0.128 & 0.179 & 0.253 & 0.419 \\
WY & 0.047 & 0.082 & 0.109 & 0.118 & 0.239 & 0.425 \\
\bottomrule
\end{tabular}
\end{table}

\end{document}